\pdfoutput=1

\documentclass[11pt]{article}

\usepackage[preprint]{acl}

\usepackage{times}
\usepackage{latexsym}

\usepackage[T1]{fontenc}

\usepackage[utf8]{inputenc}

\usepackage{microtype}

\usepackage{inconsolata}

\usepackage{graphicx}

\usepackage[normalem]{ulem}

\usepackage{graphicx}
\usepackage{booktabs}

\newcommand{\name}[0]{\textsc{WildFrame}}
\newcommand{\spike}[0]{SPIKE}
\newcommand{\llama}[1]{Llama-3#1B}
\newcommand{\gemma}[1]{Gemma-2-#1B}
\newcommand{\mixtral}[1]{Mixtral-v0.1-#1B}
\newcommand{\mistral}[0]{Mistral-v0.3-7B}
\newcommand{\gptfam}[0]{GPT}
\newcommand{\gpt}[1]{\gptfam-#1}
\newcommand{\gemini}[0]{Gemini-2.5-flash}

\usepackage{subcaption}
\usepackage{xcolor}
\definecolor{basecolor}{HTML}{8C564B}
\definecolor{framecolor}{HTML}{1F77B4}   
\usepackage{booktabs}
\usepackage{multirow}
\usepackage[normalem]{ulem} 
\usepackage{pifont}
\newcommand{\cmark}{\ding{51}} 
\newcommand{\xmark}{\ding{55}} 

\usepackage{graphicx}
\usepackage{subcaption}

\title{Comparing the Framing Effect in Humans and LLMs \\on Naturally Occurring Texts}

\author{Gili Lior \qquad  Liron Naccache \qquad Gabriel Stanovsky \\
        The Hebrew University of Jerusalem \\ \texttt{\{gili.lior,gabriel.stanovsky\}@mail.huji.ac.il}}

\begin{document}
\maketitle
\begin{abstract}
Humans are influenced by how information is presented, a phenomenon known as the \emph{framing effect}. Prior work suggests that LLMs may also be susceptible to framing, but it has relied on synthetic data and did not compare to human behavior. To address this gap, we introduce \name{} -- a dataset for evaluating LLM responses to positive and negative framing in naturally-occurring sentences, alongside human responses on the same data. \name{} consists of 1,000 real-world texts selected to convey a clear sentiment; we then \textit{reframe} each text in either a positive or negative light and collect human sentiment annotations. Evaluating eleven LLMs on \name{}, we find that all models respond to reframing in a human-like manner ($r\geq0.52$), and that both humans and models are influenced more by positive than negative reframing. Notably, \gptfam{} models are the least correlated with human behavior among all tested models. These findings raise a discussion around the goals of state-of-the-art LLM development and whether models should align closely with human behavior, to preserve cognitive phenomena such as the framing effect, or instead mitigate such biases in favor of fairness and consistency.\footnote{\name{} can be found at~\url{https://huggingface.co/datasets/gililior/WildFrame}}
\end{abstract}

\section{Introduction}
The \textit{framing effect} is a well-known cognitive phenomenon, where different presentations of the same underlying facts affect how those facts are perceived by humans~\citep{tversky1981framing}. For example, presenting an economic policy as only creating 50,000 new jobs, versus  also reporting that it would cost 2B USD, can dramatically shift public opinion ~\cite{sniderman2004structure}. 

Previous research has shown that LLMs exhibit various cognitive biases, including the framing effect~\cite{lore2024strategic,shaikh2024cbeval,malberg2024comprehensive,echterhoff-etal-2024-cognitive}. However, existing studies either rely on synthetic datasets or evaluate LLMs and humans on different inputs, limiting direct behavioral comparison. In addition, comparisons between models and humans typically use a single metric for evaluation rather than comparing patterns in human behavior. 

\begin{figure}[tb!]
    \centering
    \includegraphics[width=0.93\linewidth]{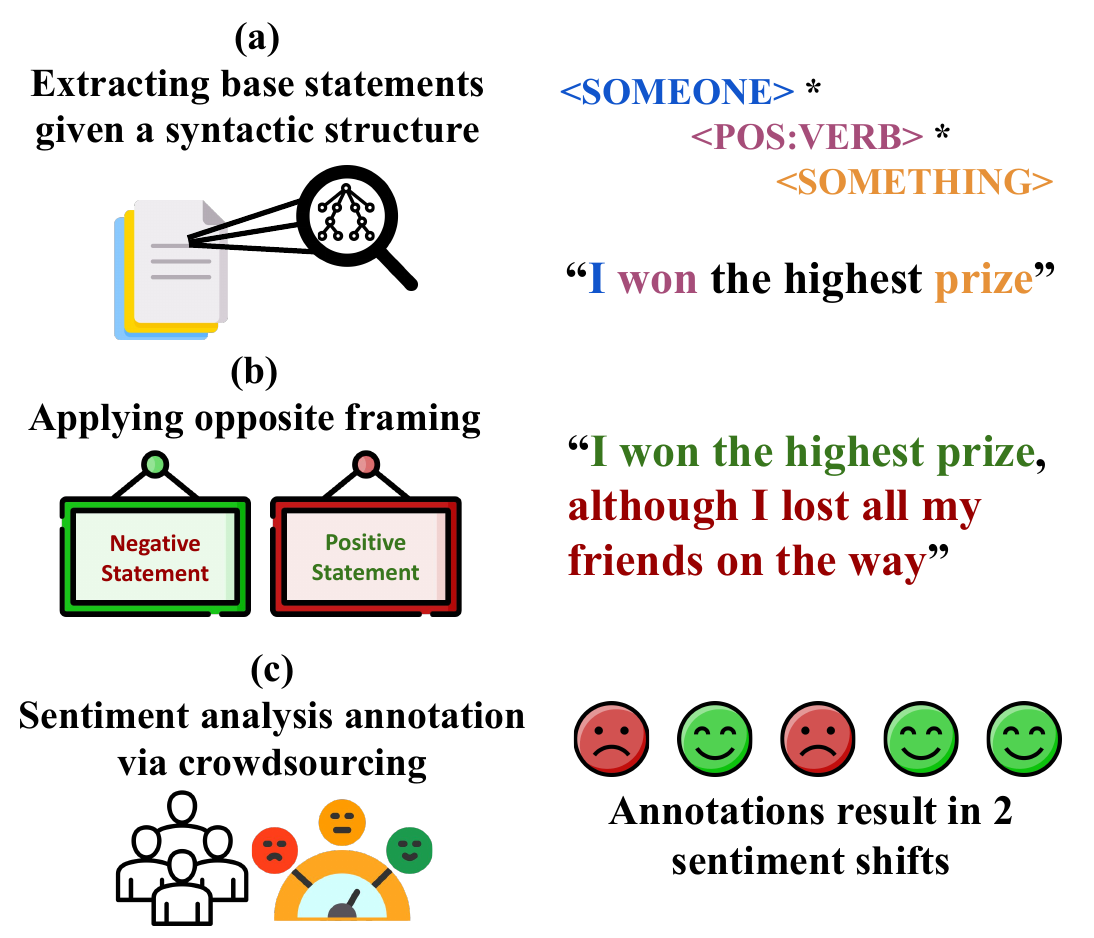}
    \caption{The \name{} data construction process. In step (a) we extract statements based on their syntactic structure, aiming for statements with clear negative or positive sentiment. Next, in (b), we \emph{reframe} the statement by adding a suffix or prefix, conveying the \emph{opposite} sentiment. Finally, in (c),  five annotators mark the sentiment of the reframed statement, counting how many annotators shift sentiment, i.e., 
    the reframed statement sentiment is opposite to the base sentiment. The red parts in the figure represent negative parts of statement, while green represents positive parts.}
    \label{fig:fig1}
\end{figure}

In this work, we evaluate LLMs on real-world data that is reframed to convey a different sentiment than the sentiment of the original statement. We adopt a broad definition of framing inspired by prior work examining evaluative framing~\cite{tong2021good}, which includes adding information that may change the reader’s perspective. For example, reframing ``\textit{I won the highest prize}'' to ``\textit{I won the highest prize, \uline{although I lost all my friends on the way}}''. 
This setup corresponds to real-world scenarios in which additional information is selectively introduced to influence interpretation. Such framing is common in everyday communication, including in political discourse and social media, where judgments often depend on which aspects of an event are emphasized rather than on the event itself. Consequently, understanding the effect that framing has on LLMs becomes increasingly important, as they are  integrated into real-world systems.

Rather than measuring model performance in terms of accuracy, we analyze how closely LLM responses align with human annotations. Furthermore, while previous studies have examined the effect of framing on decision making, we extend this analysis to sentiment analysis, as sentiment perception plays a key explanatory role in decision-making \cite{lerner2015emotion}. Studying framing in this work allows us to highlight systematic differences between humans and LLMs, which can later be informative when differentiating cases in which human-like behavior may be appropriate, versus when more rationale, framing-invariant behavior may be desirable.

To better understand the effect of different framings on LLMs in comparison to human behavior, we introduce the \name{} dataset (Section~\ref{sec:data}), comprising 1,000 statements, curated through a three-step process, as shown in Figure~\ref{fig:fig1}. First, we collect a set of real-world statements that express a clear negative or positive sentiment (e.g., ``\textit{I won the highest prize}'' vs ``\textit{I failed my math test}’’). Second, we \emph{reframe} the text by appending or prepending a clause with opposite sentiment (e.g., ``\textit{I won the highest prize, \uline{although I lost all my friends on the way}}''). Finally, we collect human annotations by asking  participants if they consider the reframed statement to be overall positive or negative.

\name{} consists of Amazon reviews, which is chosen since sentiment in review statements is more robust, compared to domains such as news, which introduce confounding factors like prior political leanings~\cite{druckman2004political}. Examples from \name{} are presented in Table~\ref{tab:after-framing}.

Recent work has raised concerns that human annotations collected via crowdsourcing platforms may be contaminated by LLM-assisted responses~\cite{veselovsky2023artificial}. This poses a key methodological challenge for studies comparing humans and LLMs: how to retain the scalability of crowdsourcing while ensuring that observed patterns reflect genuine human behavior rather than implicit model usage. In this work, we take several steps to increase confidence in the authenticity of our annotations, including validating against established cognitive science findings, statistical analyses of variance, and comparisons with a small-scale, controlled in-house human baseline.

We then evaluate eleven LLMs, including both proprietary and open-source models, on the \name{} dataset, and compare them against human annotations. Specifically, we compare the \textit{sentiment shifts} in humans and models. That is, labeling the sentiment aligning with the reframing sentiment, rather than the base sentiment, e.g., voting ``negative'' for the statement ``\textit{I won the highest prize, \uline{although I lost all my friends on the way}}'', as it aligns with the opposite framing sentiment. \emph{Sentiment shift} is also exemplified in Table~\ref{tab:after-framing}.

We find that LLMs are influenced by framing in a similar manner to human behavior, with all models exhibiting a strong correlation with human annotations ($r \geq 0.52$). Additionally, both humans and LLMs are more affected by positive reframing of negative base statements than by negative reframing of positive base statements. This asymmetry aligns with prior findings in cognitive science, which show that positive framing tends to exert a stronger influence~\cite{tong2021good, GRAPPI2024114341}. We also observe a general trend in which larger models tend to exhibit higher correlation with human behavior.

Interestingly, we find that \gpt{3.5} and \gpt{4o}~\cite{openai2024gpt4osystemcard} exhibit the lowest correlation with human behavior, and \gpt{5-mini}~\cite{openai2025gpt5intro} shows correlation levels that are comparable to several smaller open-source models. This observation raises questions about how architectural design choices and training procedures may influence a model’s susceptibility to framing effects.

This work contributes to understanding the parallels between LLM and human cognition, offering insights into how cognitive mechanisms such as the framing effect emerge in LLMs. This also raises an interesting discussion on the objective when developing state-of-the-art LLMs -- whether models should aim to align closely with human behavior, preserving cognitive phenomena such as framing effects, or instead mitigate such biases in favor of fairness and consistency. We hope that our results would benefit model developers by informing decisions on whether to harness or reduce framing effects, depending on the downstream application.

\begin{table*}[]
\resizebox{\textwidth}{!}{%
\begin{tabular}{@{}lllc@{}}
\toprule
\textbf{Reframed Sentence} &
  \textbf{Base Sentiment} &
  \textbf{Human Majority} & \textbf{Sentiment Shift} \\ \midrule
\begin{tabular}[c]{@{}l@{}}\textit{Given the price, I can't complain at all. \uline{However, the product}}\\\textit{\uline{quality is sub-par.}} \end{tabular} &
  Positive &
  Negative (0.8) & \cmark \\ \midrule
  \begin{tabular}[c]{@{}l@{}}\textit{Stanly doesn't fail to deliver with this model. \uline{It's unfortunately}}\\\textit{\uline{quite pricey for its features, though.}} \end{tabular} &
  Positive &
  Positive (0.6) & \xmark \\ \midrule
    \begin{tabular}[c]{@{}l@{}}\textit{I won't waste my money again, and hopefully you won't either,}\\\textit{\uline{but it was a valuable lesson about the importance of making}}\\\textit{\uline{wise financial choices.}}
  \end{tabular} &
  Negative &
   Negative (0.8) & \xmark \\ \midrule
  \begin{tabular}[c]{@{}l@{}}\textit{For me, Aspartame causes bad memory loss and nasty}\\\textit{gastrointestinal distress, \uline{but this has encouraged me to seek out}}\\\textit{\uline{healthier, natural  alternatives and cultivate a balanced diet.}}\end{tabular} &
  Negative &
  Positive (0.8) & \cmark \\  \bottomrule
\end{tabular}%
}
\caption{
Examples from \name{}. The underlined text indicates the reframing clause, while the remaining text corresponds to the base statement. 
The Base Sentiment column reflects the dominant sentiment of the base statement.
Human Majority Vote denotes the majority sentiment assigned to the reframed sentence, with the fraction of annotators in agreement shown in parentheses.
Sentiment Shift indicates whether the majority human judgment aligns with the reframing sentiment (\cmark) or remains aligned with the base sentiment (\xmark).
}

\label{tab:after-framing}
\end{table*}


\section{The \name{} Dataset}\label{sec:data}




Our dataset curation consists of three steps, as depicted in Figure~\ref{fig:fig1}. First, we collect natural, real-world statements, with some clear sentiment, either positive or negative (\S\ref{sec:base-statements}; e.g., ``\textit{I won the highest prize}'' as positive). Next, using an LLM, 
we reframe each statement by  appending or prepending a clause conveying the opposite sentiment
(\S\ref{sec:adding-framing}; e.g., ``\textit{I won the highest prize, \uline{although I lost all my friends on the way}}''). Finally, we collect large-scale human annotations via crowdsourcing, to label the sentiment shifts when wrapping the statements with the opposite framing (\S\ref{sec:human-annotations}; e.g., labeling ``negative'' the statement ``\textit{I won the highest prize, \uline{although I lost all my friends on the way}}'', shown also in Table~\ref{tab:after-framing}). 

The complete dataset consists of 1000 statements, in which 500 are statements that their base form has positive sentiment, and 500 are base negative statements.

\subsection{Collecting Base Statements}\label{sec:base-statements}
First, we collect base statements, which convey a clear sentiment, either clearly positive or clearly negative statements. We use \spike{} -- an extractive search system, which allows to extract statements from real-world datasets~\cite{taub-tabib-etal-2020-interactive}.
Specifically, we collect English statements from Amazon Reviews dataset, which are naturally occurring, sentiment-rich, texts but are less likely to trigger strong preexisting biases or emotional reactions, which may be a confound for our experiment.\footnote{~\url{https://spike.apps.allenai.org/datasets/reviews}} 

Using \spike, we extract ${\sim}6k$ statements that fulfilled our designated queries, which we found correlated with clear sentiment. We designed the queries to capture positive or negative verbs that describe actions with some clear sentiment (e.g., ``\textit{enjoy}'' or ``\textit{waste}''), or statements with positive or negative adjective, describing an outcome with a clear sentiment (e.g., ``\textit{good}'' or ``\textit{nasty}''). The patterns and queries used for extraction are detailed in Appendix~\ref{sec:appendix-spike}.
Next, we run in-house annotations to label and filter the extracted statements, to handle negations or other cases where the statement does not convey a clear sentiment. 
The filtering process results in $1,301$ positive statements, and $1,229$ negative statements. 

To assess annotation quality, we measured inter-annotator agreement over 300 sampled sentences. Annotators showed strong alignment in sentiment judgments: only $1\%$ of sentences exhibited strong disagreement, where annotators assigned opposite sentiment labels. In contrast, a much larger source of disagreement arose when deciding whether a sentence is suitable for the task, with $37.1\%$ of cases reflecting weak disagreement tied to this suitability criterion. We hypothesize that suitability is more subjective and influenced by individual preferences. Importantly, when considering only the sentences that both annotators agreed were suitable, sentiment agreement increases to $99\%$, demonstrating highly consistent sentiment judgments once suitability is established.

\subsection{Adding Framing}\label{sec:adding-framing}

We reframe the statements in our dataset using three different methods, generating three reframed versions for each base statement, to support robust LLM evaluation.

For the most basic paraphrasing, we use \gpt{4}~\cite{achiam2023gpt}. 
The input prompt includes a one-shot example followed by the instruction: ``Add a \texttt{SENTIMENT} suffix or prefix to the given statement. Do not change the original statement,'' where \texttt{SENTIMENT} is either ``positive'' or ``negative,'' opposite to the base statement sentiment (i.e., positive framing for a negative base statement, and vice versa).

\begin{figure}[tb!]
    \centering
    \includegraphics[width=\linewidth]{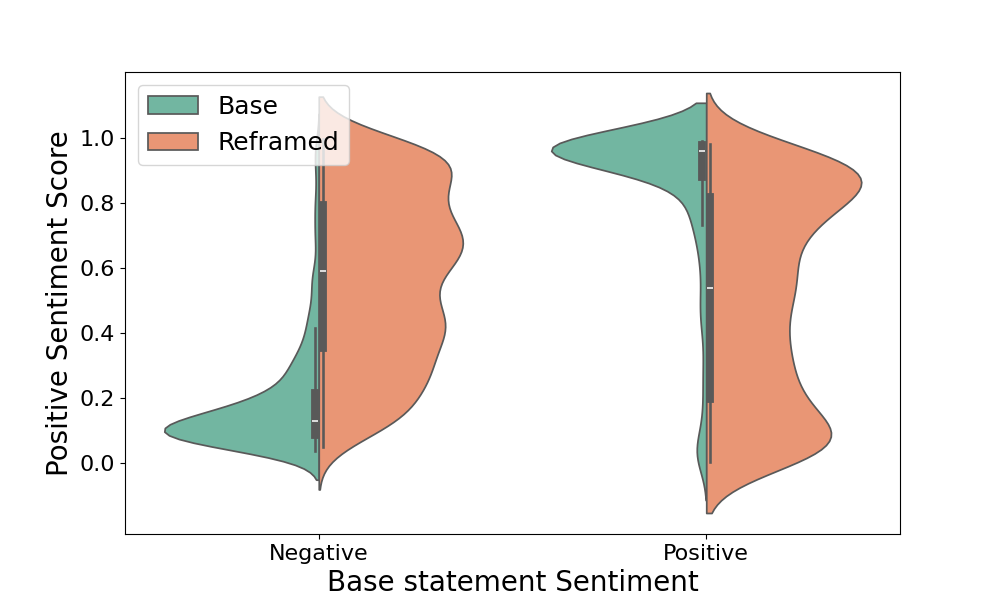}
    \caption{Distribution of sentiment scores before and after applying opposite-sentiment framing, as detailed in Section~\ref{sec:adding-framing}. Prior to framing, base sentences exhibit a clear polarity (positive or negative), whereas the application of opposite framing introduces ambiguity, shifting the sentiment scores toward a less distinct polarity.}
    \label{fig:pos-score-dist}
\end{figure}

Unlike the base statements, the sentiment conveyed by reframed statements is inherently more ambiguous, with no single clear label. This is illustrated in Figure~\ref{fig:pos-score-dist}, by the distribution of sentiment scores before and after reframing, given by a fine-tuned sentiment analysis model.\footnote{~\url{https://huggingface.co/cardiffnlp/twitter-roberta-base-sentiment-latest}}
This ambiguity enables us to measure the extent to which LLMs shift their sentiment predictions following framing, and how closely shifts align with human behavior.

Due to the known LLM sensitivity to changes in the prompt~\cite{sclar2023quantifying,mizrahi-etal-2024-state}, to strengthen the reliability of LLM evaluation, we generate two additional reframings for each sentence, and test LLMs on them. The full prompts are provided in Appendix~\ref{sec:framing-prompts}.


\subsection{Collecting Human Annotations}\label{sec:human-annotations}

In the final step, we collect human annotations through Amazon Mechanical Turk to evaluate the framing effect in \name{} over human participants.\footnote{\url{https://www.mturk.com}} 
Details about the annotation platform are elaborated in Appendix~\ref{sec:mturk-appendix}.

The complete dataset includes 1K statements, reframed by \gpt{4} as explained in Section~\ref{sec:adding-framing}, each annotated by five different annotators. Collecting five annotations per statement enables robust estimation of statement ambiguity and inter-annotator agreement.

For the annotation process, each statement in our dataset is presented in its reframed version (i.e., positive base statements with negative framing and vice versa), to five different annotators. This setup generates, for each dataset instance, a \textit{sentiment shift} score ranging from 0 to 5, representing the number of annotators that votes for the sentiment that aligns with the reframing, which means that the overall sentiment of the reframed statement has shifted from its base sentiment. For example, in Figure~\ref{fig:fig1}, the statement ``\textit{I won the highest prize, \uline{although I lost all my friends on the way}}'' is shown to have two annotators voting ``negative'', which aligns with the sentiment of the framing and not the base statement, so the label for that instance in \name{} would be 2 (sentiment shifts).

Instances with score near 0 indicate that annotators agree that the overall sentiment remains unchanged despite the opposite sentiment reframing. Score closer to 5 indicates that annotators agree that reframing shifted the perceived sentiment, while score around 2-3 suggests that the reframing makes the sentiment ambiguous.

\section{Results}\label{sec:results}

We evaluate eleven LLMs on the \name{} dataset, comprising both open- and closed-source instruction-tuned models. These included \gpt{3.5}, \gpt{4o} and \gpt{5-mini}~\cite{openai2025gpt5intro}, Gemini-2.5-flash~\cite{comanici2025gemini}, \llama{.1-8} and \llama{.3-70}~\cite{dubey2024llama}, \mistral{}~\cite{jiang2023mistral}, \mixtral{8x7} and \mixtral{8x22}~\cite{mistral2023mixtral}, \gemma{9} and \gemma{27}~\cite{team2024gemma}.

\subsection{Validation of Actual Human Behavior in Crowdsourced Annotations}

Recent discussions have raised concerns that annotations obtained via crowdsourcing platforms like Mechanical Turk might not reflect genuine human input, as workers could potentially use LLM-generated responses~\cite{veselovsky2023artificial}. While in-house controlled experiments guarantee authenticity, they typically lack the scale required for robust LLM evaluation. This raises a methodological challenge for research comparing humans and model behavior: how can we maintain the scalability benefits of crowdsourcing while ensuring we are testing against real human signal? In particular, it is difficult to distinguish between cases where humans and LLMs truly behave similarly and cases where apparent similarity arises because annotators themselves rely on LLMs.


In this study, we assess annotators' authenticity using three complementary components: alignment with prior cognitive science findings, statistical analysis of variance within annotators and within model predictions, and a comparison against a controlled in-house human baseline.

First, we verify that the crowdsourced annotations reproduce well-established cognitive phenomena. If the annotations failed to exhibit documented cognitive phenomena, this would suggest that crowdsourcing is an ineffective proxy for human behavior. Indeed, our finding that humans are more affected by positive framing than by negative framing is consistent with prior findings in cognitive science~\cite{10.2307/2118508,tong2021good, GRAPPI2024114341}.

\begin{figure}
    \centering
    \includegraphics[width=\linewidth]{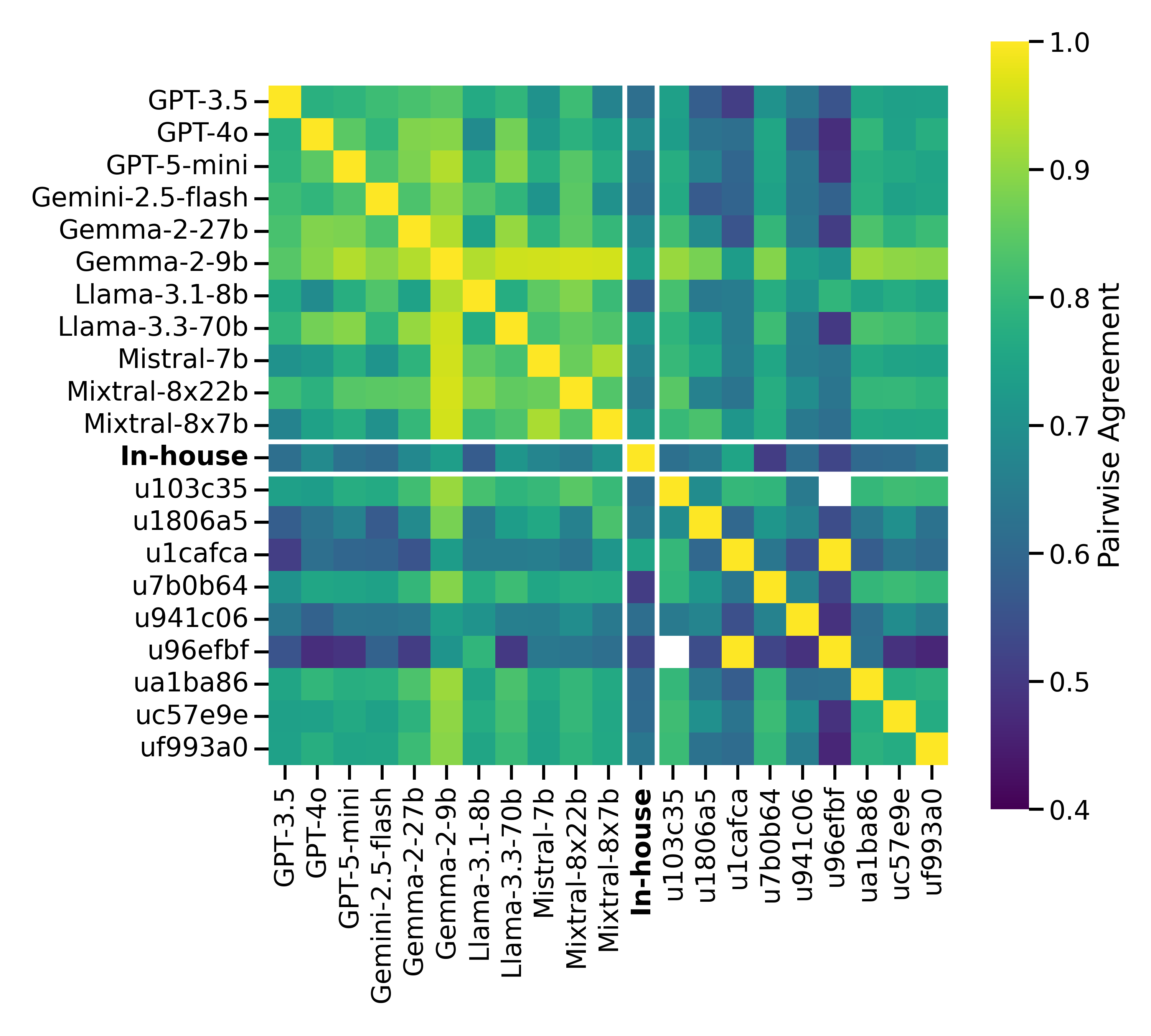}
    \caption{Pairwise agreement rates between LLMs and human annotators. The matrix is divided into model-to-model (top-left), human-to-human (bottom-right), and cross-comparisons. The ``In-house'' row represents labels from controlled graduate student annotators, serving as a quality anchor. The visual density shows that the model cluster is significantly tighter (higher agreement) than the human cluster, which displays the natural variance expected in subjective tasks.}
    \label{fig:all-agreement}
\end{figure}

Second, as shown in Figure~\ref{fig:all-agreement}, no single annotator exhibits perfect agreement with any individual LLM which were out at the time of annotations. However, since it is infeasible to test every available model, relying on individual pairwise agreement is insufficient. Instead, we analyze the collective behavior of the models, treating them as a representative LLM prediction cluster. As indicated by the bright yellow block in the upper-left of Figure~\ref{fig:all-agreement}, this cluster is significantly tighter (lower variance) than the human annotator bottom-right block. To quantify this, we compared the variance of human annotators to the variance of the LLM cluster. Previous research has established that human subjective judgments typically exhibit high variance~\cite{plank-2022-problem, sap-etal-2022-annotators}. Indeed, our Mechanical Turk annotators show a variance of $0.137$, while the LLMs show a lower variance of $0.108$. This disparity suggests that the crowd workers are operating with a degree of subjectivity characteristic of humans, rather than the mode-collapsed uniformity of models.

Finally, to enable a finer-grained analysis of individual annotator behavior, we conducted a small-scale controlled experiment with in-house annotators, consisting of eight graduate students. Each annotator labeled a distinct subset of statements, resulting in 160 annotations collected under controlled conditions without access to LLM assistance.
For this analysis, we integrated the in-house annotations with the crowdsourced labels to form a human consensus (Turkers + In-House). We then measured, for each individual annotator, whether their labels aligned more frequently with the human consensus (excluding their own vote) or with the LLM cluster, as reported in Table~\ref{tab:turker_affinity} in the Appendix.
Interestingly, the in-house baseline exhibits a slight preference toward the LLM cluster over the human consensus (1.95\% diff). Relative to this baseline, all Turkers but one align more closely with the human consensus than the in-house annotators do. The single Turker showing higher agreement with the LLM cluster (2.99\% diff) nevertheless exhibits high agreement with both groups (${\sim}80\%$). This can be used to mark this annotator as ``suspicious'', though it remains unclear how to interpret this.

To sum up, as the availability of LLMs grows, such validation strategies to verify human annotations will become essential for maintaining the reliability of large-scale human-LLM comparisons.

\begin{figure}
    \centering
    \includegraphics[width=\linewidth]{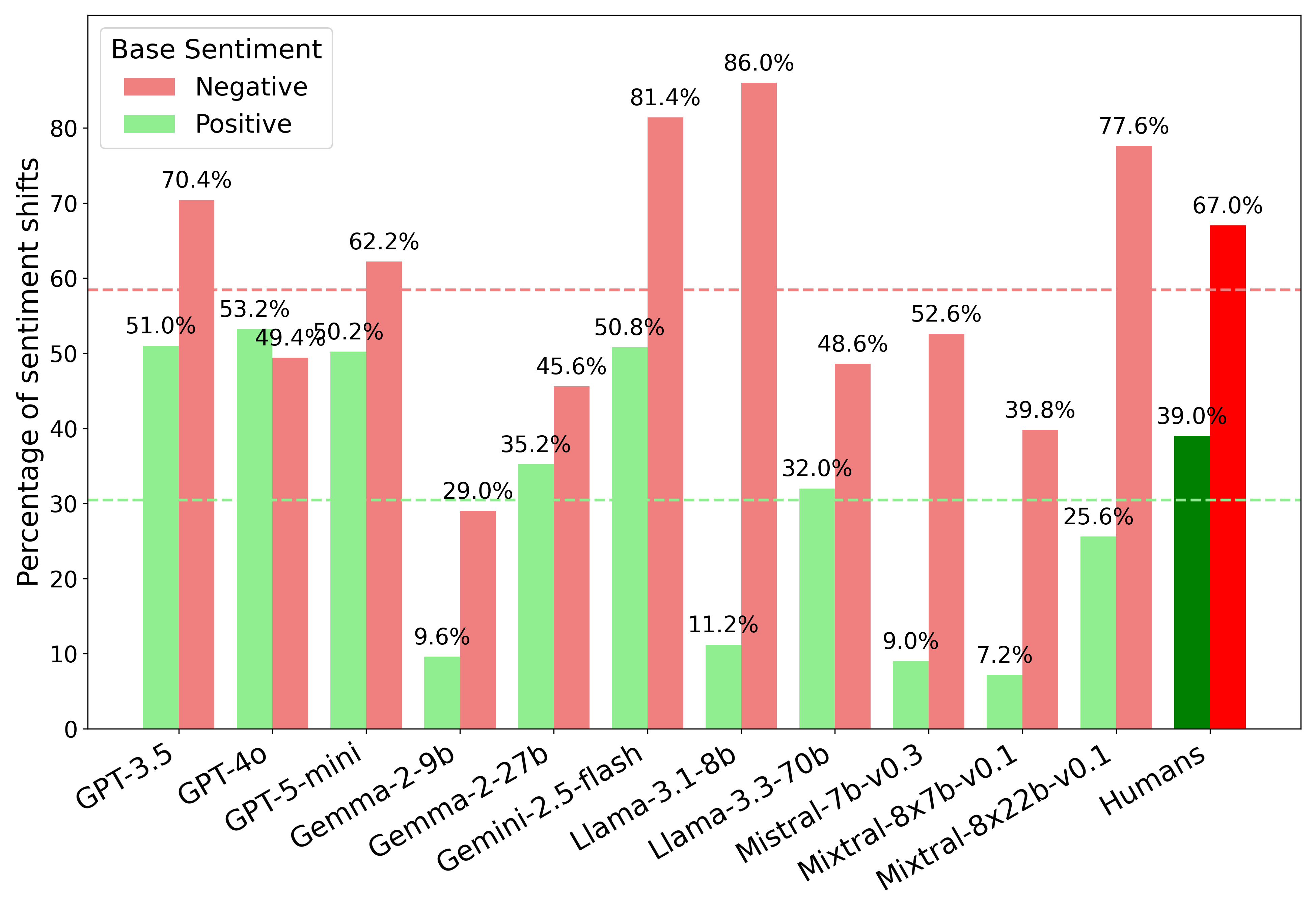}
    \caption{Percentage of reframed statements that results in sentiment shift positive to negative (or vice versa). Red represents negative-base statements reframed as positive, and green represents positive-base statements reframed as negative. Horizontal lines show the mean across models. We find that both LLMs and humans are influenced by opposite framing, with a stronger effect for positive reframing.}
    \label{fig:models-sentiment-shifts}
\end{figure}

\subsection{Analysis}

Below we identify the key observations from our results. The LLM analysis presented below is based on a single prompt design, and in Figure~\ref{fig:appendix-sentiment-shifts} in Appendix we show that the reported trends remain consistent across different prompt variations.

\paragraph{Humans and LLMs are largely more influenced by positive framing applied to negative statements than by negative framing applied to positive statements.} 
This trend is evident in Figure~\ref{fig:models-sentiment-shifts}. For LLMs, all models except \gpt{4o} exhibit higher ratios of sentiment shifts for statements that were originally negative (red bars) compared to those that were originally positive (green bars). For humans, we count the cases in which the majority voted for a sentiment shift (3 annotators or more). 
This finding aligns with~\citet{tong2021good}, which demonstrated that positive framing is more effective on humans when spatial distance is minimal. In \name{}, all statements are presented in a first-person perspective (e.g., ``\textit{I won the highest prize}''), creating zero spatial distance from the annotator's perception and thereby amplifying the impact of positive framing. 
As LLMs are increasingly deployed in real-world applications, this framing bias carries dual implications. On the one hand, it could be beneficially leveraged in supportive domains like education or mental health, where encouraging framing can enhance user motivation. On the other hand, in high-stakes settings such as legal advice, financial forecasting, or hiring, this sensitivity raises concerns regarding fairness and consistency, as model outputs may fluctuate based on superficial phrasing rather than substantive content.

\begin{figure}[t!]
    \centering
    \includegraphics[width=\linewidth]{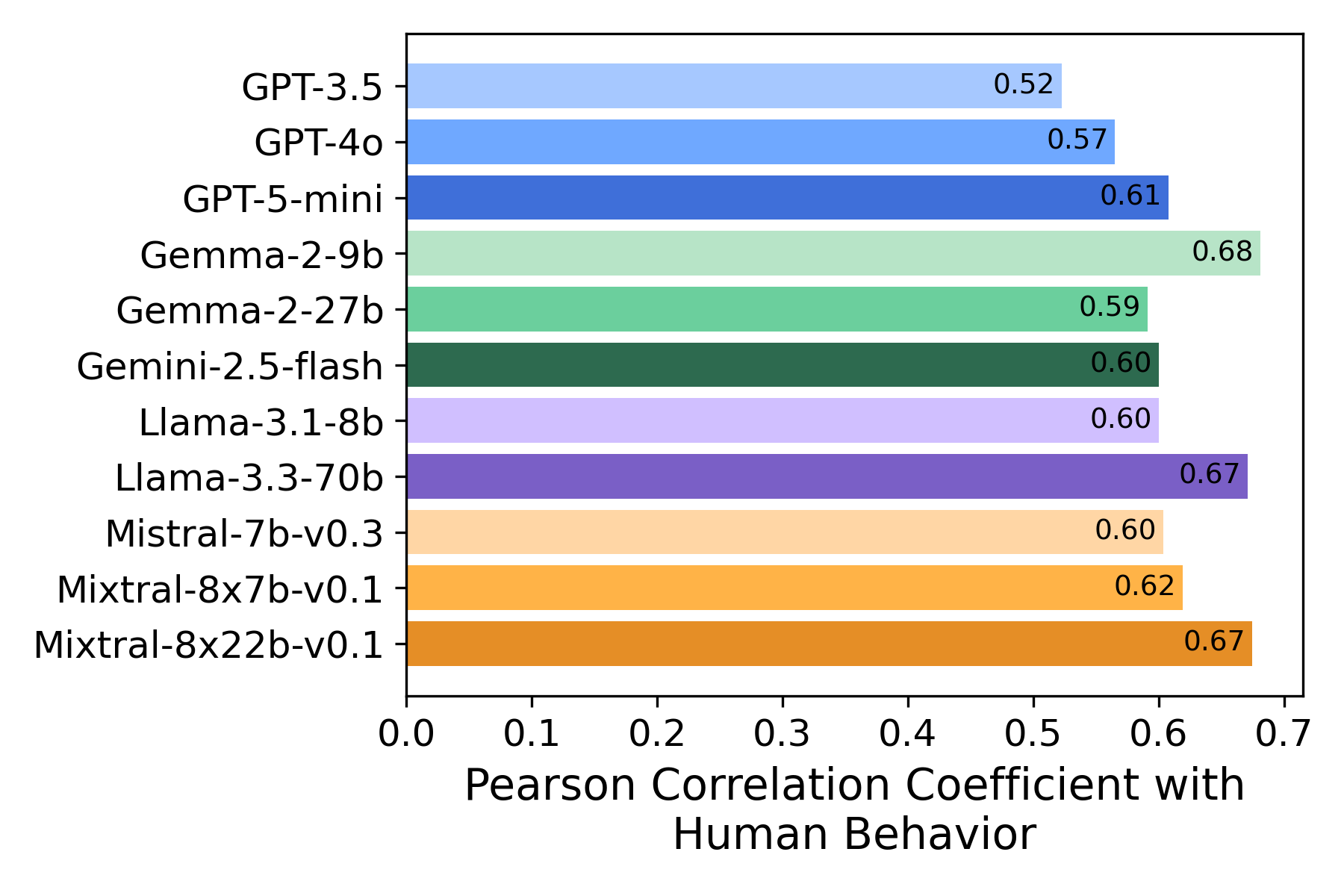}
    \caption{Pearson correlation coefficients between human sentiment shifts and predictions from various LLMs after applying opposite sentiment framing. Higher values indicate stronger alignment between the model's behavior and human annotations.}
    \label{fig:correlation-to-humans}
\end{figure}

\paragraph{Model size partially correlates with similarity to human behavior under opposite framings.}
In Figure~\ref{fig:correlation-to-humans}, we present the correlation between each model's behavior and human responses.\footnote{Human responses are soft scores of the percentage of annotators voted for negative/positive sentiment. Models could vote ``neutral''. Hence, we report the correlation between these two soft scores.} For the Llama and Mistral model families, we observe a trend where larger models with more parameters exhibit higher correlation with human behavior regarding the framing effect. However, the Gemma family shows the opposite trend. This discrepancy highlights the need for further investigation into the relationship between model size, the framing effect, and its alignment with human behavior.

\paragraph{\gptfam{} models exhibit relatively low correlation with human behavior among all other tested models.}
As shown in Figure~\ref{fig:correlation-to-humans}, both \gpt{3.5} and \gpt{4o} exhibit the weakest correlation with human behavior among all tested models, and \gpt{5-mini} reaches correlation levels comparable to some relatively small open-source models. This result is notable given the strong downstream performance typically associated with \gptfam{} models, and suggests that high benchmark performance does not necessarily entail close alignment with human responses to framing. While we cannot analyze the underlying causes of these differences due to the proprietary nature of these models, this finding raises a broader question regarding what users perceive as a best-performing model: one that most closely mirrors human behavior, or one that optimizes benchmark performance. We argue that the answer is task-dependent. In domains such as education or mental health, human-like sensitivity may be desirable, whereas in more objective settings, such as legal or regulatory applications, robustness to cognitive biases like framing may be preferable.

\paragraph{\gptfam{} models increasingly align with human framing behavior.}
As shown in Figure~\ref{fig:correlation-to-humans}, successive \gptfam{} model versions exhibit increasing correlation with human behavior under framing. \gpt{3.5} shows the lowest correlation with humans ($r=0.52$), followed by \gpt{4o} ($r=0.57$), while the most recent \gpt{5-mini} achieves the highest correlation ($r=0.61$). This trend suggests that newer models behave more similarly to humans in their response to framing, despite this behavior not being directly assessed by standard instruction-following or downstream task benchmarks. This finding is particularly noteworthy given the growing reliance on synthetic data in model training, which might be expected to increase model-model similarity rather than model-human similarity. At the same time, alignment-oriented training such as RLHF may counterbalance this effect by implicitly encouraging human-like judgment patterns. While the proprietary nature of these models prevents definitive conclusions about the underlying causes, the observed trend nonetheless points to an emergent convergence between newer \gptfam{} models and human framing behavior.

\begin{figure}[t!]
    \centering
    \includegraphics[width=\linewidth]{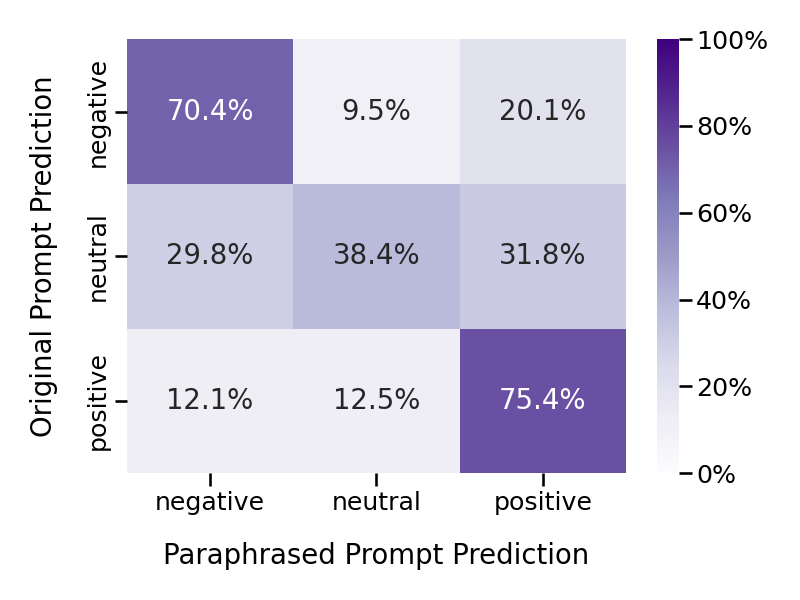}
    \caption{Aggregated consistency of LLM sentiment predictions under prompt paraphrasing. The heatmap displays a row-normalized confusion matrix, where each row represents the sentiment assigned to the original prompt, and the columns show the distribution of labels for the paraphrased version. Detailed breakdowns for individual models are provided in Figure~\ref{fig:eleven_subfigures} in Appendix.}
\label{fig:agg_consistency}
\end{figure}

\paragraph{Model sentiment predictions are robust to paraphrasing in $\sim$70\% of cases.} 
Figure~\ref{fig:agg_consistency} depicts the consistency of model predictions under \textit{different paraphrasing of the same facts}. We observe that for sentences originally labeled as positive or negative, over 70\% maintain their label. The remaining cases are split between shifting to a neutral label and flipping to the opposite sentiment. This indicates that while LLM sensitivity to phrasing introduces some noise, the overall trends reported in our analysis remain consistent across prompt variations.

\paragraph{Different kinds of reframing alter individual predictions but preserve the overall trend.} When we take the same base sentence and reframe it with different information (e.g., ``\textit{\uline{Although I lost my friends on the way}, I won the highest prize}'' vs. ``\textit{\uline{Despite the scandal ruining my reputation,} I won the highest prize}''), we find that instance-level sentiment predictions change in approximately 55\% of cases. However, the trend of the models' higher susceptibility to positive framing remains consistent, as shown in Figure~\ref{fig:appendix-sentiment-shifts} in Appendix.

\subsection{Qualitative analysis of disagreements between humans and \gpt{5-mini}}

\begin{figure}
    \centering
    \includegraphics[width=0.75\linewidth]{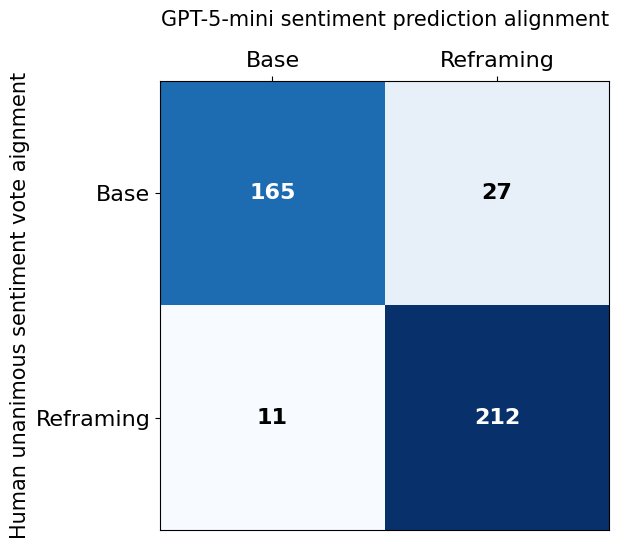}
    \caption{Counts of cases with unanimous human annotations. The figure shows the distribution of agreements and disagreements between \gpt{5-mini} predictions (columns) and unanimous human votes (rows). ``Base'' denotes predictions or votes that align with the sentiment of the original statement, while ``Reframing'' denotes predictions or votes that align with the sentiment introduced by the reframing clause.}
    \label{fig:counts}
\end{figure}

As \gpt{5-mini} is the most recent model among those we evalaute, we conduct a qualitative analysis of its disagreements with human annotations. We focus on unanimous cases, where all annotators agree on the sentiment, and the model predicts a clear sentiment (i.e., did not predict ``neutral'').
In Figure~\ref{fig:counts} we present the distribution of these cases. 

We manually go over these agreement and disagreement cases, and try to characterize these cases. Below we elaborate our findings. In all of the exmaples, the reframing part is \uline{underlined}.

Across the data, it seems that humans and models agree in following the reframing sentiment when the base statement sentiment is relatively weak or hedged, and the reframing clause contains a clear evaluative signal.
For example, in ``\textit{It doesn't dissolve smoothly, there is always some that you have to work on to get dissolved. \uline{Nevertheless, it's an excellent way to be more attentive and meticulous in preparation.}}'', 
or in ``\textit{I doubt it'll ever leave the box, but I'm sure it is really well made. \uline{Unfortunately, it’s just not as useful as I had hoped.}}'', both humans and models align with the reframing. Agreement from both humans and \gpt{5-mini} to follow the reframing also arises when it provides a salient reappraisal of the base statement, such as ``\textit{I can't blame her for being so cold, her past was hell; \uline{nevertheless, it's admirable how she remains resilient and strong despite it all.}}''.

Conversely, both humans and models align with the base statement sentiment when it expresses a categorical or final judgment that the reframing does not override. For instance, in ``\textit{I hate this, and will not buy it again}'' both sides preserve the original negative sentiment despite the subsequent silver lining, ``\textit{\uline{On the bright side, this gives me a chance to explore other options and find something that may suit me better}}''.

Disagreements arise in more ambiguous boundary cases. When the model shifts toward the reframing sentiment but humans unanimously align with the base statement, the first clause typically expresses strong positive affect (e.g., ``\textit{I love it}'', ``\textit{we love them}''), while the reframing introduces a concrete but arguably secondary complaint. For example, in ``\textit{I love it! \uline{But it’s too expensive}}'' and ``\textit{We love this pot. \uline{It’s just too bad it’s a bit too small for our needs}}'' humans tend to preserve the overall positive sentiment, treating the complaint as a caveat, whereas the model shifts toward the reframing clause. This suggests that the model may overweight explicit post-marker  content even when humans interpret it as a minor qualification.

In contrast, when humans unanimously shift toward the reframing sentiment but the model aligns with the base statement, the first clause often contains severe or emotionally charged negativity, while the reframing introduces an abstract reappraisal framed in terms of growth, resilience, or opportunity. For example, in ``\textit{Abuse in childhood causes terrible trauma... \uline{However, it also provides a chance for resilience and growth}}'' or ``\textit{Witches hate children and kill them in this story,  \uline{but this serves as the catalyst for bravery and resourcefulness}}'', humans frequently align with the reframed interpretation, while the model remains anchored to the initial negative content.

Overall, these patterns may indicate that disagreements between humans and \gpt{5-mini} are not arbitrary, but arise systematically from how each weighs affective strength, pragmatic cost, and abstract reappraisal when integrating multi-clause ambiguous statements.


\section{Discussion}
We introduced \name{}, a dataset designed to enable direct comparison between human and LLM responses to evaluative framing on naturally occurring text. Across all evaluated models, we observe consistent sensitivity to framing that correlates with human judgments, including a shared asymmetry whereby positive reframing has a stronger influence than negative reframing. This pattern mirrors well-established findings in cognitive science and supports the validity of both the dataset construction and the collected human annotations.

Crucially, we do not treat higher human-model similarity as an objective in itself. Rather, \name{} provides a diagnostic lens for characterizing how models integrate additional evaluative information when multiple perspectives are presented. The observed differences across model families, particularly the relatively lower correlation of some state-of-the-art proprietary models such as \gptfam, suggest that strong benchmark performance does not necessarily imply human-like sensitivity to framing.

More broadly, our findings highlight framing sensitivity as a task-dependent property. In interactive or supportive settings, human-like responsiveness may improve usability and engagement, while in high-stakes or normative contexts, robustness to framing may be preferable. \name{} offers a controlled benchmark for studying this trade-off and for informing future model design and evaluation.

\section{Limitations}
In this work, we address a cognitive bias, and as with any research involving human participants, our study has several limitations.

First, our framing experiment is conducted within a single domain -- Amazon reviews -- and focuses on a specific type of statement. Some of our findings may be artifacts of this dataset rather than generalizable patterns.


Furthermore, our study focuses solely on sentiment analysis. Other downstream tasks influenced by framing, such as question answering or decision-making, may exhibit different patterns of sensitivity. Investigating these tasks could provide further insights into the broader impact of framing on LLM behavior in real-world applications.


\bibliography{costum}

\appendix



\section{Extracting data with \spike}\label{sec:appendix-spike}

We found two patterns of statements, which can convey a clear sentiment, and built queries upon these patterns to extract statements from \spike. Examples for all types of statements are presented in Table~\ref{tab:base-sentence}.

First, are statements in which the verb in the statements is a verb with clear sentiment, that often implies the sentiment of the entire statement. E.g., `wastes', `rejects', `fails' are negative verbs, while verbs like `enjoys', `succeeds', `empowers', conveys positive statements. 

The second pattern of statements that we found suitable for conveying a clear sentiment, are statements which describe some event/action, and its consequences, where often the adjective that describes the consequences holds information whether it is positive or negative. 

Next, we needed to label and filter them due to two main issues. First, we needed to handle the cases in which negation words appear in the statement and flips the sentiment. For example, a statement like ``We did not enjoy the show'' includes a positive verb (enjoy), but the negation flips its sentiment to be a negative statement. Another issue we encountered is that there are many statements which are irrelevant to our case, even though they match the positive/negative patterns, for example ``I couldn't sympathize with the shopping aspect of the book since I hate to shop .'' does not convey any clear sentiment, despite the use of the verb `hate'.

\begin{table*}[t]
\centering
\resizebox{\textwidth}{!}{
\begin{tabular}{ll}
\toprule
\textbf{Category} & \textbf{Example Sentence} \\
\midrule
Positive Verb & ``To my surprise I did \textbf{enjoy} the book and the characters .'' \\
Negative Verb & ``This dock has done nothing but provide frustration and \textbf{waste} a great deal of my time trying to get it to work properly .'' \\
\midrule
Positive Outcome & ``This bag provides \textbf{good} protection for my snare drum at a really \textbf{good} price .'' \\
Negative Outcome & ``For me , Aspartame causes \textbf{bad} memory loss and \textbf{nasty} gastrointestinal distress .'' \\
\bottomrule
\end{tabular}
}
\caption{Examples for base statements collected using \spike. The words that inflect the sentiment are in bold.}
\label{tab:base-sentence}

\end{table*}

\subsection{SPIKE Queries}
\begin{enumerate}
    \item :something :[{pos/neg verbs}]develops
    \item:something :[{pos/neg adjectives}]badly :[{cause synonym}]causes :something
\end{enumerate}

\subsection{Word Lists}

\paragraph{Positive verbs.} achieve, admire, affirm, appreciate, aspire, awe, bless, blossom, celebrate, cherish, comfort, contribute, delight, donate, elevate, empower, enchant, encourage, energize, engage, enjoy, enrich, enthuse, excel, fervor, flourish, fortify, glisten, glow, gratitude, grow, harmonize, heal, illuminate, innovate, inspire, invigorate, laugh, learn, liberate, love, motivate, nourish, nurture, praise, prosper, radiate, rally, refresh, rejoice, renew, revel, revere, revitalize, savor, shine, smile, soar, spark, sparkle, stimulate, strengthen, succeed, support, synergize, thrive, unite, uplift, volunteer, adore, amaze, boost, captivate, win.

\paragraph{Negative verbs.} abandon, abuse, accuse, alienate, begrudge, betray, bewilder, blame, collapse, complain, condemn, confuse, contradict, criticize, decay, deceive, decline, defeat, demoralize, deny, despair, destroy, deteriorate, devalue, discourage, discriminate, dishearten, dismantle, dismiss, dissolve, doubt, exploit, fail, falter, fear, frustrate, grieve, harass, hate, hurt, ignore, inhibit, intimidate, lose, mock, overlook, overwhelm, pollute, punish, regress, reject, repress, resent, sabotage, shatter, sicken, stifle, suffer, suffocate, suppress, terrorize, torment, undermine, violate, waste, weaken, whine, withdraw, withhold, worry.

\paragraph{Positive adjectives.}
admirable, lucky, enjoyable, magnificent, enthusiastic, marvelous, euphoric, amazing, excellent, exceptional, amused, excited, amusing, extraordinary, nice, noble, outstanding, appreciative, fabulous, overjoyed, astonishing, fantastic, benevolent, fortunate, pleasant, blissful, pleasurable, brilliant, positive, glad, prominent, good, proud, charming, cheerful, reliable, gracious, grateful, clever, great, happy, superb, superior, terrific, incredible, tremendous, inspirational, delighted, delightful, joyful, joyous, uplifting, wonderful, lovely.

\paragraph{Negative adjectives.}
sad, angry, upset, disgusting, boring, disappointing, frustrating, annoying, miserable, terrible, deppressing, unhappy, melancolic, heartbreaking, Furious, iritating, emberessing, horrible, stupid, unlucky, negative, bad.

\paragraph{``Causes'' synonym.} causes, creates, generates, prompts, produces, induces, yields, affects, invokes, effectuates, results, encourages, promotes, introduces, begets, engenders, occasions, develops, starts, contributes, initiates, inaugurates, establishes, begins, cultivates, acquires, provides, launches.

\section{Adding Framing}\label{sec:framing-prompts}

We used three different methods to generate the reframings.

\subsection{Reframing with \gpt{4}}
These reframings were shown to human annotators and are the main reframings we analyze in this paper.

We used the \texttt{gpt-4-0613} version for generating the reframing, using the following prompts:

\begin{itemize}
    \item ``Here is an example of a base statement with a negative sentiment: I failed my math test today. Here is the same statement, after adding a positive framing: I failed my math test today, however I see it as an opportunity to learn and improve in the future. Here is a negative statement: \texttt{STATEMENT} Like the example, add a positive suffix or prefix to it. Don't change the original statement.''

    \item ``Here is an example of a base statement with a positive sentiment: I got an A on my math test. Here is the same statement, after adding a negative framing: I got an A on my math test. I think I spent too much time learning it though. Here is a positive statement: \texttt{STATEMENT}. Like the example, add a negative suffix or prefix to it. Don't change the original statement.''
\end{itemize}

\subsection{Paraphrasing the reframings with \gemini}

Next, we utilized \gemini{}~\cite{comanici2025gemini} to paraphrase the already-reframed sentences, while preserving factual content but varying style, order, and wording.

We used the following prompt:

\begin{itemize}
    \item ``Given this original sentence: \texttt{STATEMENT}, a former process has generated a reframing sentence. Your task is to rewrite the reframing, while maintaining its meaning and keep the same facts, but reframe it in a different way. Rewrite the following sentence using a different style or structure (e.g., change the tone, word order, or phrasing) while strictly preserving the original meaning and factual details.\\Original Sentence: \texttt{STATEMENT}.\\Original Reframing Sentence: \texttt{REFRAMED}.\\New Reframing:''
\end{itemize}

\subsection{Paraphrasing given only the base statement}

Finally, we want to generate a completly new reframing, that would introduce different information than in the first reframing. For that, we prompted \gemini{} with only the base statement, and the cognitive definition of framing. We used the following prompt:

\begin{itemize}
    \item ``In psychology and communication, Framing refers to the way information is presented to an  audience. Specifically, Attribute Framing involves highlighting different characteristics  of a single event to shift the perceived value from positive to negative (or vice versa),  while the underlying facts of the event remain objectively true. Your task is to rewrite the provided \texttt{BASE\_SENTIMENT} statement to apply \texttt{OPPOSITE\_SENTIMENT} sentiment frame, i.e., \texttt{FRAMING\_PROMPT}. You have full creative freedom to restructure the sentence, change the vocabulary, or alter the syntax, as long as you preserve the core factual event. Do not invent new facts,  but you may interpret the existing facts through a new lens.\\ Statement: \texttt{STATEMENT}\\Re-framed sentence:''
    \item Positive \texttt{FRAMING\_PROMPT}: ``rewrite it to emphasize the positive implications (opportunity, safety, freedom, or long-term benefit)''
    \item Negative \texttt{FRAMING\_PROMPT}: ``rewrite it to emphasize the negative implications (responsibility, risk, cost, or pressure).''
\end{itemize}

\section{Annotation Platform}\label{sec:mturk-appendix}

We select a pool of 10 qualified workers who successfully passed our qualification test, which consisted of 20 base statements (unframed), for which annotators were expected to achieve perfect accuracy. The estimated hourly wage for the entire experiment was approximately 14USD per hour.

\begin{figure*}
    \centering
    \includegraphics[width=\linewidth]{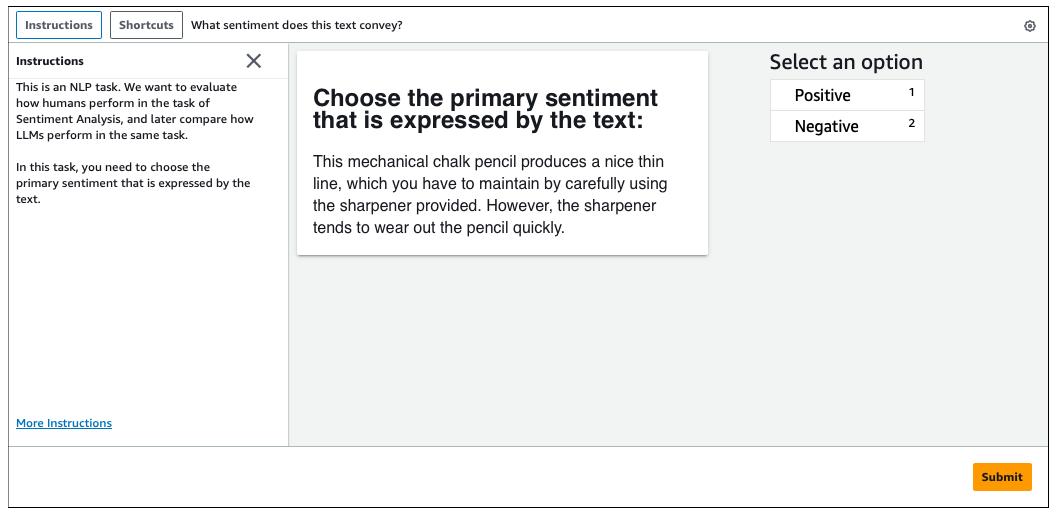}
    \caption{Screenshot of the Mechanical Turk annotation platform.}
    \label{fig:annotation-platform}
\end{figure*}

Screenshot of the annotation platform is presented in Figure~\ref{fig:annotation-platform}.

\section{Models}\label{sec:appendix-models}

We ran the open models via together-ai API.\footnote{\url{https://www.together.ai}} 
The list of models we used are:
\begin{itemize}
    \item "google/gemma-2-9b-it"
    \item "google/gemma-2-27b-it"
    \item "mistralai/Mistral-7B-Instruct-v0.3"
    \item "mistralai/Mixtral-8x7B-Instruct-v0.1"
    \item "mistralai/Mixtral-8x22B-Instruct-v0.1"
    \item "meta-llama/Llama-3.1-8B-Instruct-Turbo"
    \item "meta-llama/Llama-3.3-70B-Instruct-Turbo"
\end{itemize}

For \gptfam{} model family, we used the OpenAI api, with ``gpt-3.5-turbo-0125'', ``gpt-4o-2024-08-06'', and ``gpt-5-mini''.\footnote{\url{https://platform.openai.com/docs/overview}}

For \gemini, we used Google AI Studio API, calling ``gemini-2.5-flash''.\footnote{\url{https://aistudio.google.com}}




\section{In-House Annotations}

We used a streamlit environment to run the in-house annotation exxperiment. We had 8 participants, all are grad students that agreed to participate in our experiment without payment. 

\begin{figure*}
    \centering
    \includegraphics[width=0.7\linewidth]{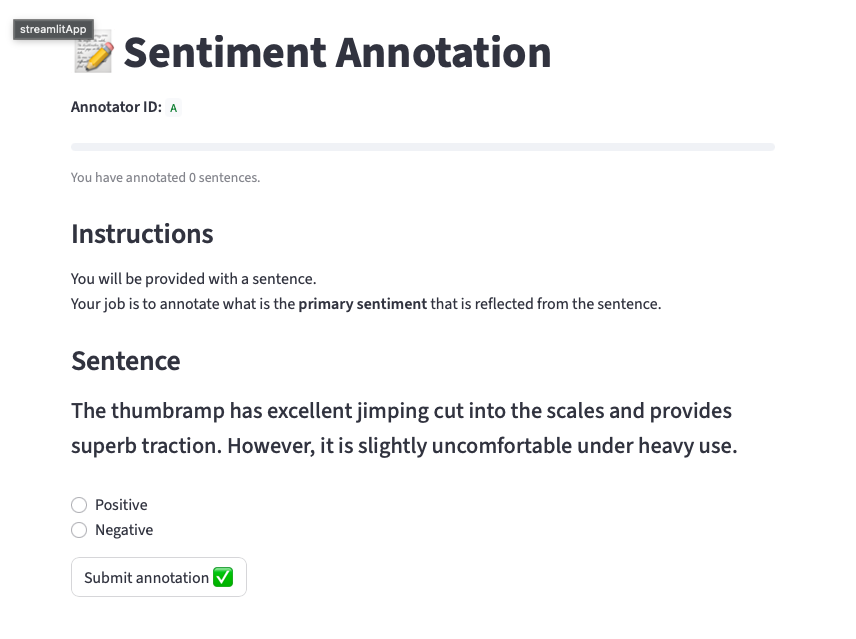}
    \caption{Screenshot of the in-house annotation platform via streamlit.}
    \label{fig:in-house-annotation-platform}
\end{figure*}

In Figure~\ref{fig:in-house-annotation-platform} we provide a screenshot of the annotation platform, and their instructions for the task were similar to the Mechanical Turk experiment.

Each annotator has annotated random unique sentences from \name{}, about 20 sentences each, resulting in a total of 160 in-house annotations which we know did not involve LLM usage.

\section{Dataset License and Usage Terms}
\label{sec:dataset_license}

The \name{} dataset is released to support academic research on framing effects in humans and large language models. The dataset is distributed under a \textbf{Creative Commons Attribution 4.0 International (CC BY 4.0)} license.

Under this license, users are permitted to share, copy, redistribute, and adapt the dataset for any purpose, including commercial use, provided that appropriate credit is given to the original authors. Attribution should include a citation to this paper and a reference to the dataset name, \textsc{\name{}}.

\paragraph{Source Data and Derivative Content}
The \name{} dataset is constructed from short excerpts of the Amazon Reviews dataset, which is publicly available and widely used for research purposes. Base statements were extracted using syntactic patterns and further filtered through human annotation. Reframed variants were generated using a large language model and therefore constitute derived textual content rather than verbatim reproductions of the original reviews.

The dataset does not include user identifiers, personal data, or metadata that could be used to identify individuals. All examples are presented as standalone sentences without links to original reviews or authors.

\paragraph{Intended Use and Limitations}
The dataset is intended for research and evaluation purposes, including the study of framing effects, sentiment ambiguity, and human--model behavioral alignment. While the CC BY 4.0 license permits broad reuse, users are responsible for ensuring that their downstream applications comply with applicable laws, ethical guidelines, and the terms of any third-party models used in conjunction with the dataset.

We encourage responsible use of \textsc{\name{}}, particularly in sensitive application domains, and recommend clearly documenting any modifications or extensions made to the dataset in derivative works.

\begin{table*}[h]
    \centering
    \begin{tabular}{lcccc}
\toprule
\textbf{Annotator ID} & \textbf{Human \%} & \textbf{LLM \%} & \textbf{Diff (Human $-$ LLM)} & \textbf{\# Annotations} \\
\midrule
\texttt{u103c35} & 75.56\% & 76.67\% & -1.11\% & 90 \\
\texttt{u1806a5} & 71.64\% & 64.18\% & +7.46\% & 67 \\
\texttt{u1cafca} & 64.52\% & 58.06\% & +6.45\% & 31 \\
\texttt{u7b0b64} & 72.88\% & 67.80\% & +5.08\% & 59 \\
\texttt{u941c06} & 67.12\% & 68.49\% & -1.37\% & 73 \\
\texttt{u96efbf} & 60.00\% & 57.14\% & +2.86\% & 35 \\
\texttt{ua1ba86} & 79.85\% & 82.84\% & -2.99\% & 134 \\
\texttt{uc57e9e} & 81.10\% & 72.44\% & +8.66\% & 127 \\
\texttt{uf993a0} & 80.52\% & 80.52\% & +0.00\% & 154 \\
\midrule
\textit{In-House} & 66.23\% & 68.18\% & -1.95\% & 154 \\
\bottomrule

    \end{tabular}
    \caption{Individual annotator agreement with human consensus (calculated via leave-one-out, including the in-house annotation) versus LLM Consensus. Negative values in the Diff column indicate higher agreement with the LLM cluster, while positive values indicate higher agreement with the human consensus. The annotation counts represent the subset of the 160 in-house control items labeled by each worker, excluding instances where the LLM ensemble produced a tie.}
    \label{tab:turker_affinity}
\end{table*}

\begin{figure*}[htbp]
    \centering
    \begin{subfigure}{0.49\textwidth} 
        \centering
        \includegraphics[width=\textwidth]{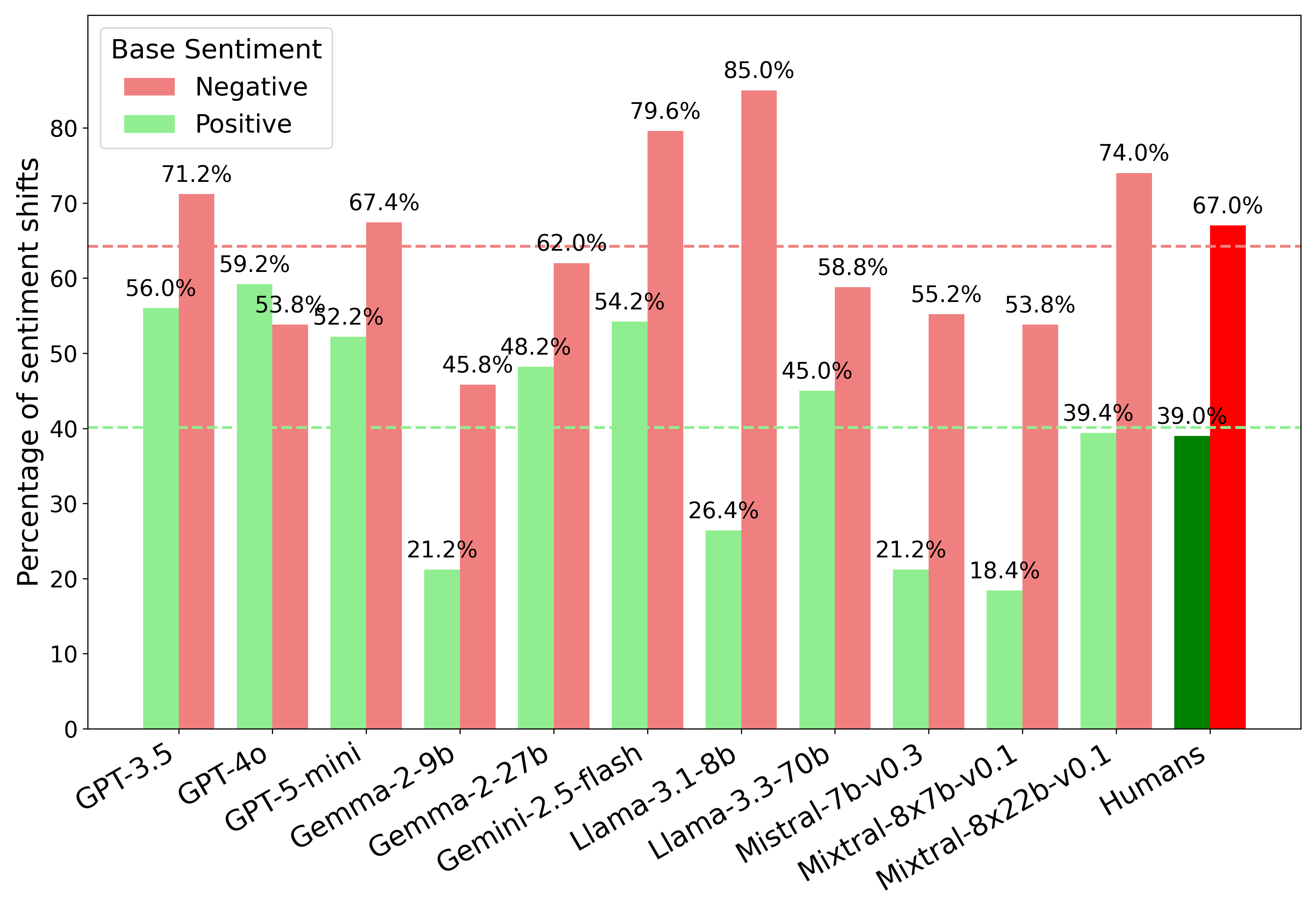} 
        \caption{Paraphrasing the reframing.}
        \label{fig:negative-flip}
    \end{subfigure}
    \begin{subfigure}{0.49\textwidth}
        \centering
        \includegraphics[width=\textwidth]{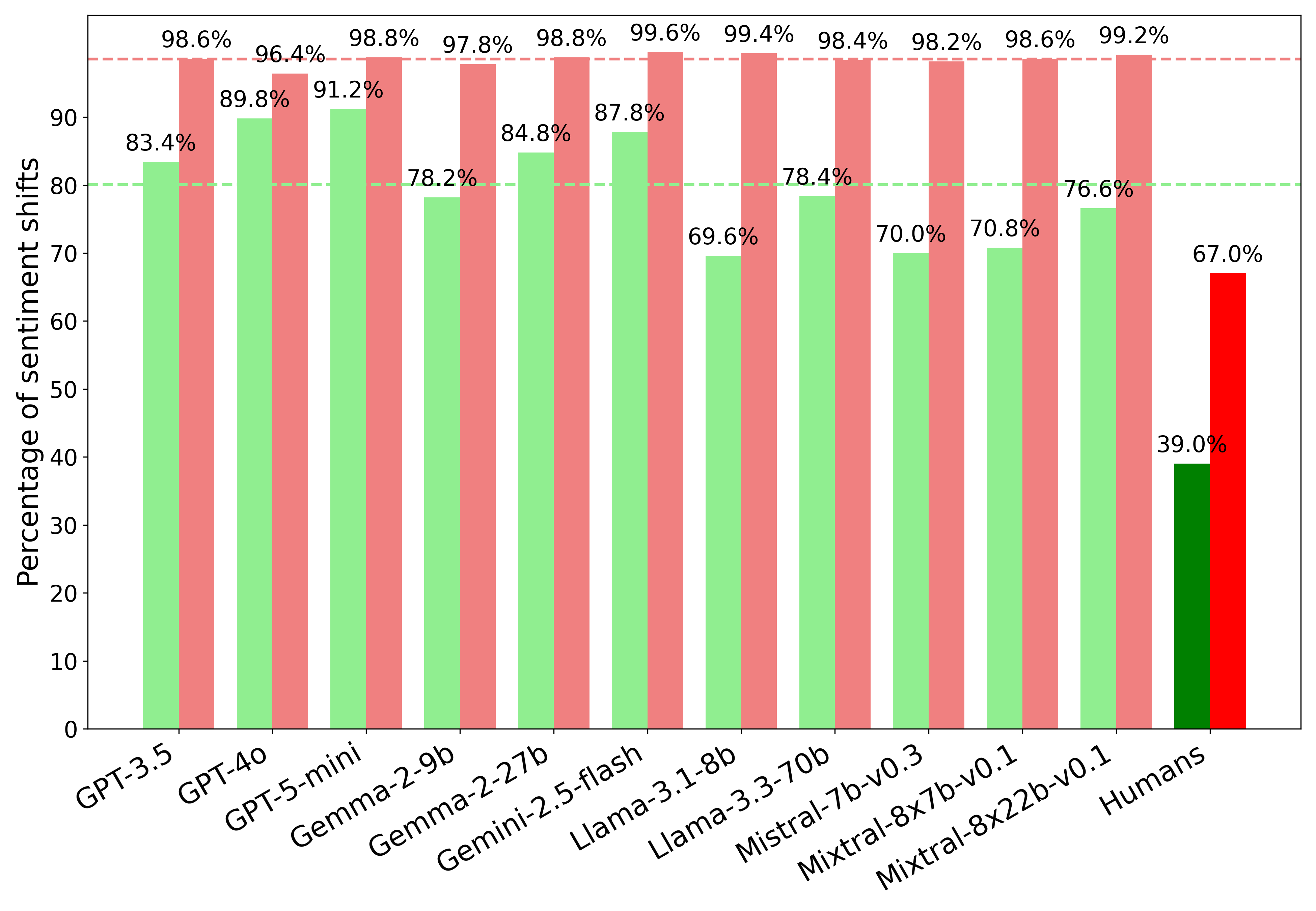} 
        \caption{Generating a new reframing from scratch.}
        \label{fig:positive-flip}
    \end{subfigure}
    \caption{Similar to Figure~\ref{fig:models-sentiment-shifts}, percentage of reframed statements that results in sentiment shift positive to negative or vice versa), on two additional reframing paraphrases.
    }
    \label{fig:appendix-sentiment-shifts}
\end{figure*}

\begin{figure*}[hb]
    \centering
    \begin{subfigure}{0.49\textwidth} 
        \centering
        \includegraphics[width=\textwidth]{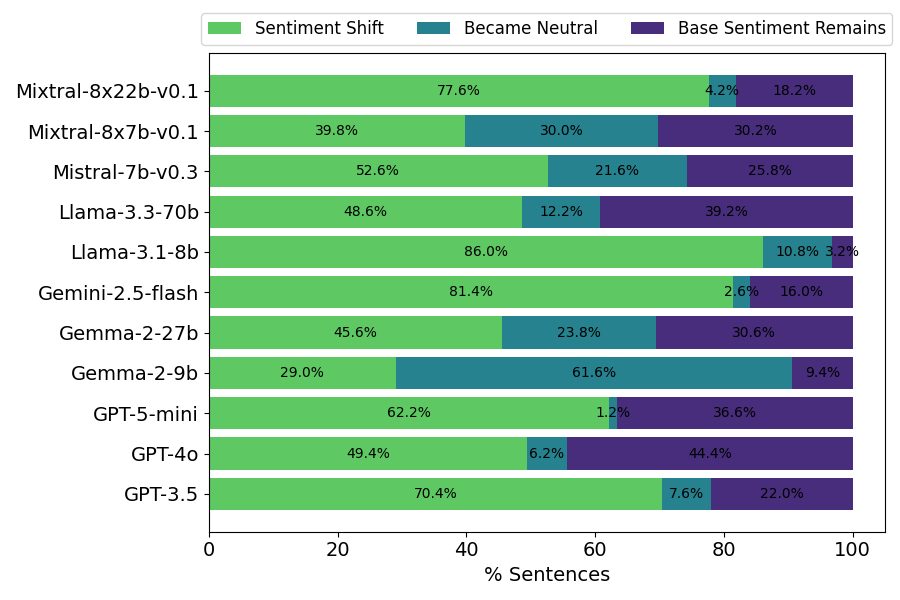} 
        \caption{Sentences that are \textbf{negative} in their original form.}
        \label{fig:negative-flip}
    \end{subfigure}
    \begin{subfigure}{0.49\textwidth}
        \centering
        \includegraphics[width=\textwidth]{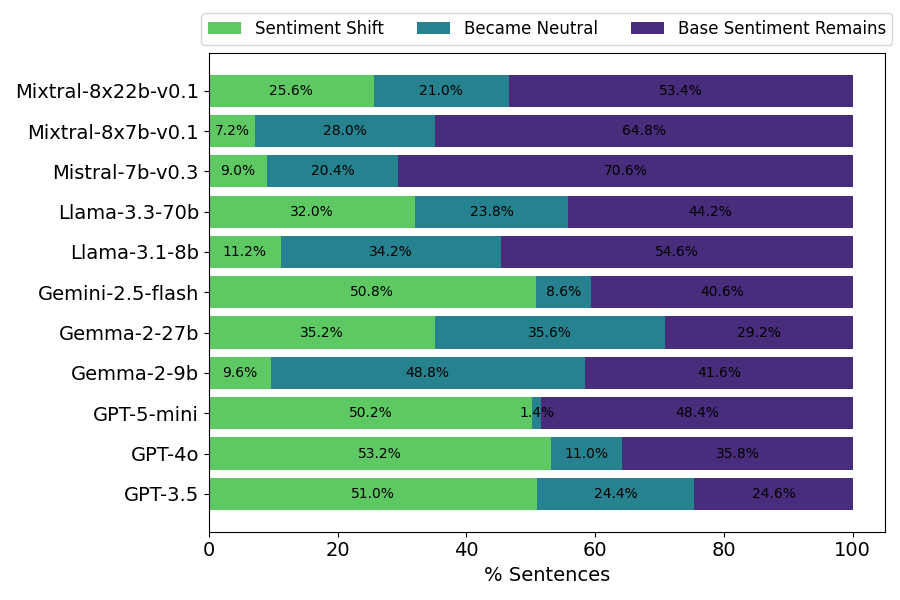} 
        \caption{Sentences that are \textbf{positive} in their original form.}
        \label{fig:positive-flip}
    \end{subfigure}
    \caption{Proportion of sentences for which LLMs flipped sentiment, became neutral, or retained the original sentiment when presented with opposite sentiment framing. For example, this measures the percentage of sentences originally labeled as positive, that were labeled as negative after applying negative framing (and vice versa).
    }
    \label{fig:flip-proportion}
\end{figure*}

\begin{figure*}[t]
    \centering

    \begin{subfigure}[t]{0.32\textwidth}
        \centering
        \includegraphics[width=\linewidth]{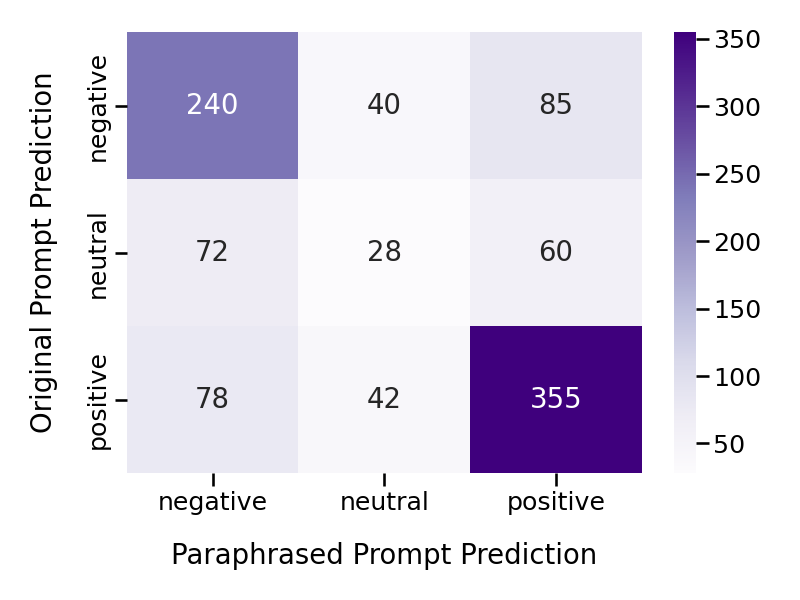}
        \caption{\gpt{3.5}}
    \end{subfigure}\hfill
    \begin{subfigure}[t]{0.32\textwidth}
        \centering
        \includegraphics[width=\linewidth]{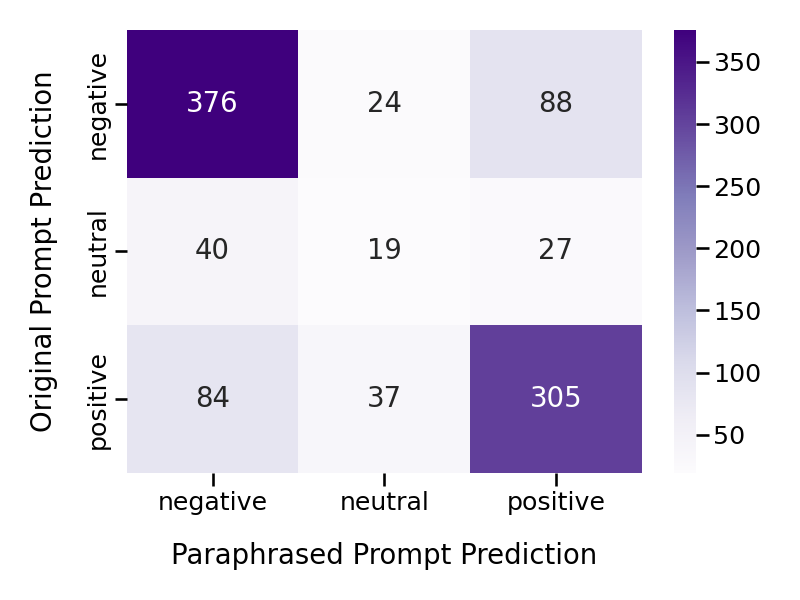}
        \caption{\gpt{4o}}
    \end{subfigure}\hfill
    \begin{subfigure}[t]{0.32\textwidth}
        \centering
        \includegraphics[width=\linewidth]{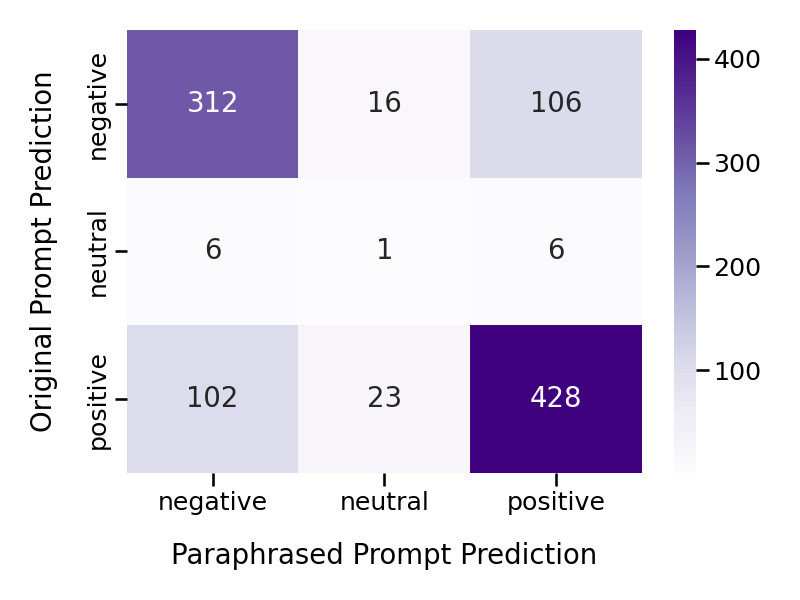}
        \caption{\gpt{5-mini}}
    \end{subfigure}

    \vspace{0.6em}

    \begin{subfigure}[t]{0.32\textwidth}
        \centering
        \includegraphics[width=\linewidth]{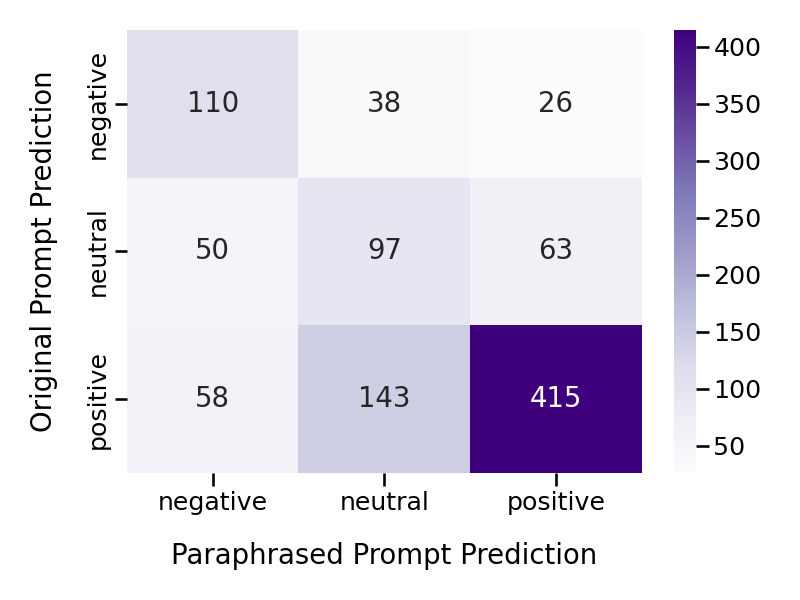}
        \caption{\mistral}
    \end{subfigure}\hfill
    \begin{subfigure}[t]{0.32\textwidth}
        \centering
        \includegraphics[width=\linewidth]{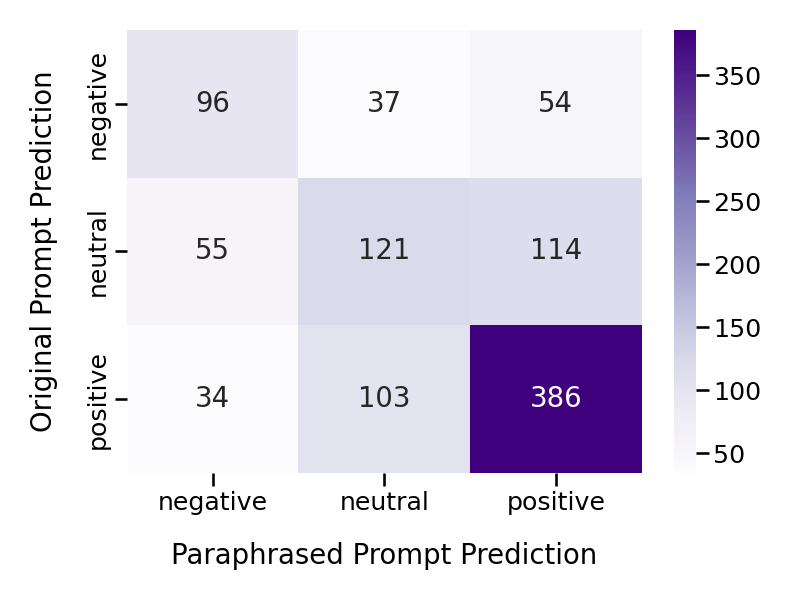}
        \caption{\mixtral{8x7}}
    \end{subfigure}\hfill
    \begin{subfigure}[t]{0.32\textwidth}
        \centering
        \includegraphics[width=\linewidth]{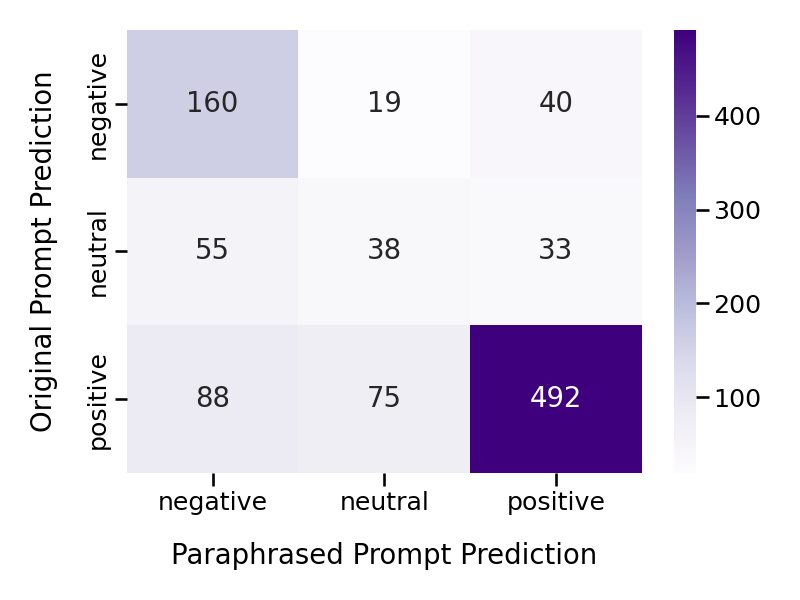}
        \caption{\mixtral{8x22}}
    \end{subfigure}

    \vspace{0.6em}

    \begin{subfigure}[t]{0.32\textwidth}
        \centering
        \includegraphics[width=\linewidth]{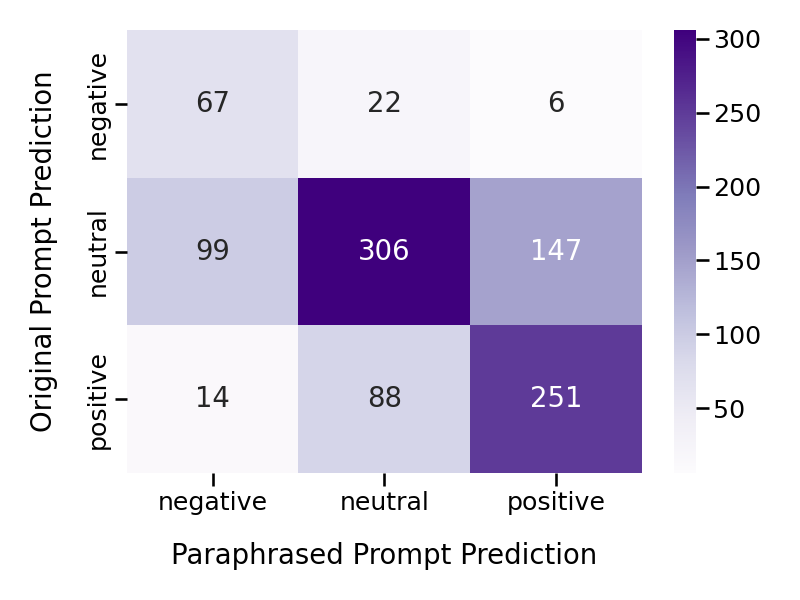}
        \caption{\gemma{9}}
    \end{subfigure}\hfill
    \begin{subfigure}[t]{0.32\textwidth}
        \centering
        \includegraphics[width=\linewidth]{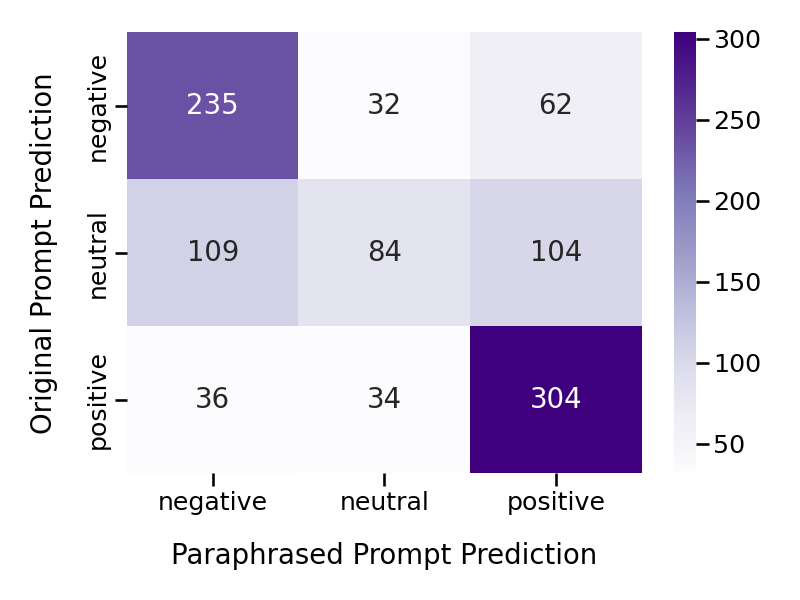}
        \caption{\gemma{27}}
    \end{subfigure}\hfill
    \begin{subfigure}[t]{0.32\textwidth}
        \centering
        \includegraphics[width=\linewidth]{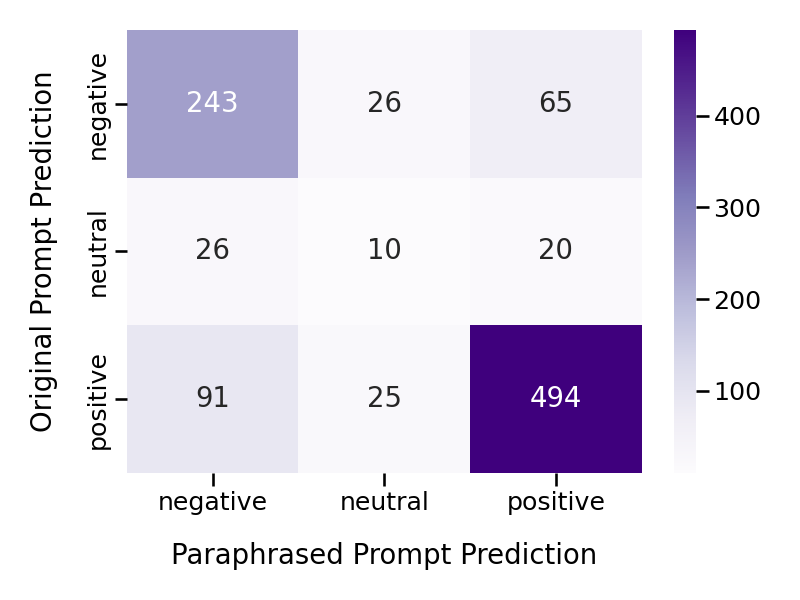}
        \caption{\gemini{}}
    \end{subfigure}

    \vspace{0.6em}

    \begin{subfigure}[t]{0.32\textwidth}
        \centering
        \includegraphics[width=\linewidth]{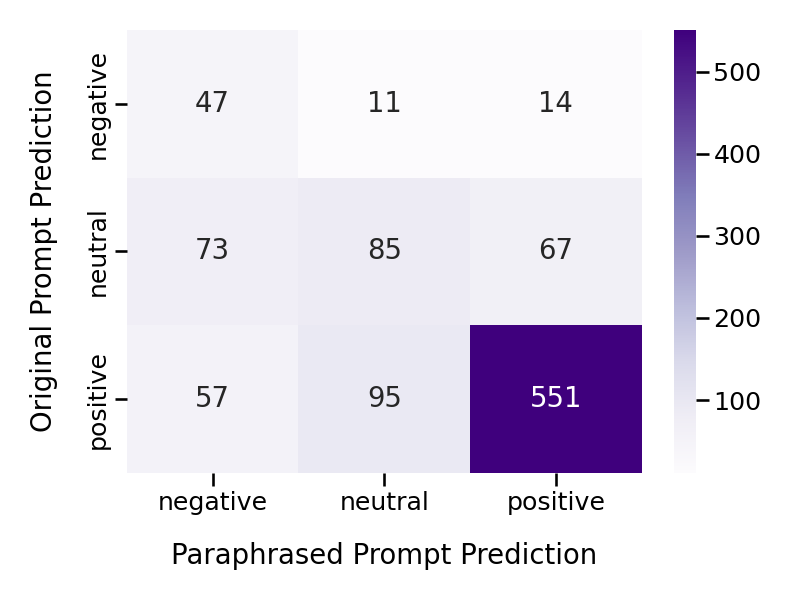}
        \caption{\llama{.1-8}}
    \end{subfigure}\hfill
    \begin{subfigure}[t]{0.32\textwidth}
        \centering
        \includegraphics[width=\linewidth]{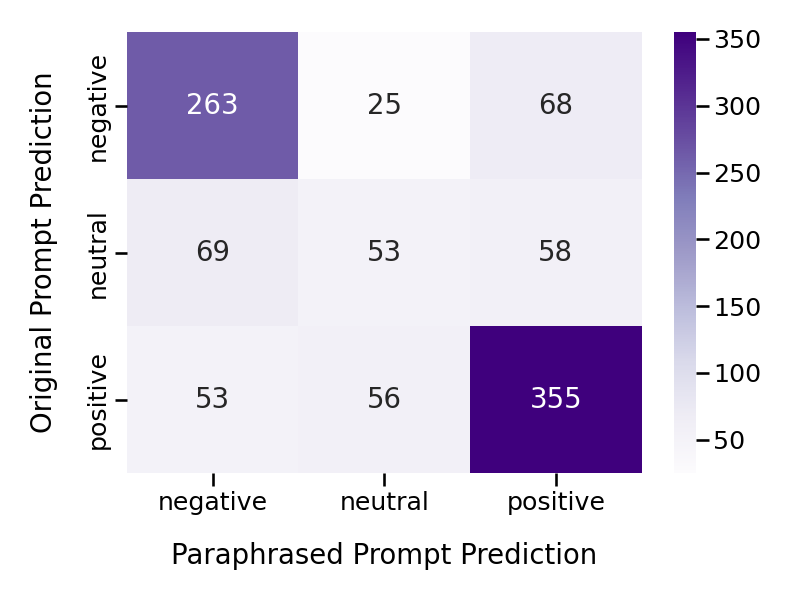}
        \caption{\llama{.3-70}}
    \end{subfigure}
    \caption{ Consistency under prompt paraphrasing per model, row normalized, corresponds to Figure~\ref{fig:agg_consistency}.}
    \label{fig:eleven_subfigures}
\end{figure*}

\begin{figure}
    \centering
    \includegraphics[width=\linewidth]{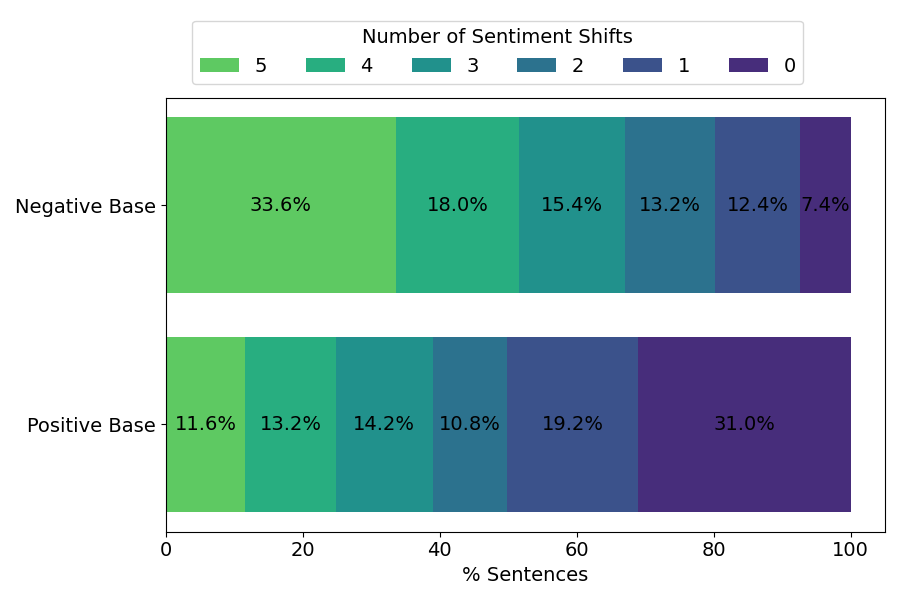}
    \caption{Proportions of sentences where annotators agreed on the extent of sentiment shift after applying opposite sentiment framing. The bars represent the percentage of sentences with 0 to 5 annotators agreeing on a sentiment shift.}
    \label{fig:humans-flip}
\end{figure}

\end{document}